\documentclass[letterpaper]{article} % DO NOT CHANGE THIS
\usepackage{aaai23}  % DO NOT CHANGE THIS
\usepackage{times}  % DO NOT CHANGE THIS
\usepackage{helvet}  % DO NOT CHANGE THIS
\usepackage{courier}  % DO NOT CHANGE THIS
\usepackage[hyphens]{url}  % DO NOT CHANGE THIS
\usepackage{graphicx} % DO NOT CHANGE THIS
\usepackage{natbib}  % DO NOT CHANGE THIS AND DO NOT ADD ANY OPTIONS TO IT
\usepackage{caption} % DO NOT CHANGE THIS AND DO NOT ADD ANY OPTIONS TO IT
\usepackage{algorithm}
\usepackage{algorithmic}
\usepackage{amsmath}
\usepackage{soul}
\usepackage{color, xcolor}
\usepackage{multirow}
\usepackage{siunitx}
\usepackage{newfloat}
\usepackage{listings}
\DeclareCaptionStyle{ruled}{labelfont=normalfont,labelsep=colon,strut=off} % DO NOT CHANGE THIS
\floatstyle{ruled}
\newfloat{listing}{tb}{lst}{}
\floatname{listing}{Listing}
\usepackage{bm}
\usepackage{amsfonts}
\usepackage{booktabs}
\usepackage[export]{adjustbox} 

\title{RLogist: Fast Observation Strategy on Whole-slide Images with \\ Deep Reinforcement Learning}
\author {
    Boxuan Zhao\textsuperscript{\rm 1, \rm 2} \footnote{Equal contribution; \dag Corresponding author}, 
    Jun Zhang\textsuperscript{\rm 1} \footnotemark[1], 
    Deheng Ye\textsuperscript{\rm 1 \dag}, 
    Jian Cao\textsuperscript{\rm 2}, 
    Xiao Han\textsuperscript{\rm 1}, 
    Qiang Fu\textsuperscript{\rm 1}, 
    Wei Yang\textsuperscript{\rm 1}
}
\affiliations {
    \textsuperscript{\rm 1} Tencent AI Lab  \\\textsuperscript{\rm 2} Shanghai Jiao Tong University\\
    b-x-zhao@sjtu.edu.cn, junejzhang@tencent.com, dericye@tencent.com, cao-jian@sjtu.edu.cn, \\ haroldhan@tencent.com, leonfu@tencent.com, willyang@tencent.com
}

\usepackage{bibentry}
\begin{document}

\maketitle

\begin{abstract}
Whole-slide images (WSI) in computational pathology have high resolution with gigapixel size, but are generally with sparse regions of interest, which leads to weak diagnostic relevance and data inefficiency for each area in the slide. 
Most of the existing methods rely on a multiple instance learning framework that requires densely sampling local patches at high magnification. 
The limitation is evident in the application stage as the heavy computation for extracting patch-level features is inevitable. 
In this paper, we develop RLogist, a benchmarking deep reinforcement learning (DRL) method for fast observation strategy on WSIs. 
Imitating the diagnostic logic of human pathologists, our RL agent learns how to find regions of observation value and obtain representative features across multiple resolution levels, without having to analyze each part of the WSI at the high magnification.
We benchmark our method on two whole-slide level classification tasks, including detection of metastases in WSIs of lymph node sections, and subtyping of lung cancer. 
Experimental results demonstrate that RLogist achieves competitive classification performance compared to typical multiple instance learning algorithms, while having a significantly short observation path. 
In addition, the observation path given by RLogist provides good decision-making interpretability, and its ability of reading path navigation can potentially be used by pathologists for educational/assistive purposes. Our code is available at: \url{https://github.com/tencent-ailab/RLogist}.
\end{abstract}

\section{Introduction}
\begin{figure}[tbp]
    \centering
    \includegraphics[width=8cm]{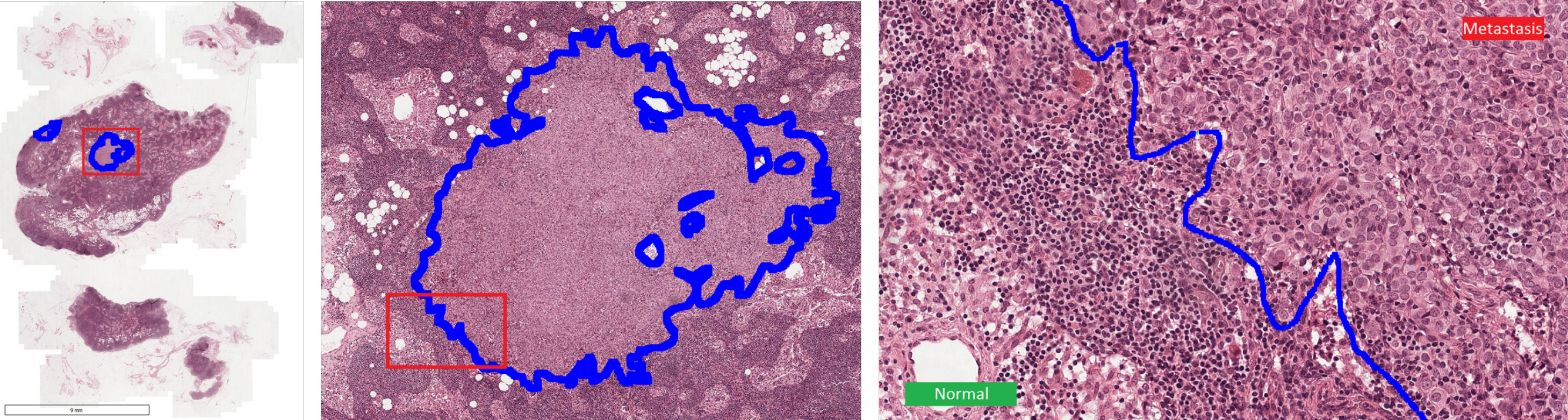}
    \caption{Metastatic and normal regions in H\&E stained WSIs of lymph node sections. In most datasets, disease-positive tissue sections take only a small portion of the whole slide and pixel-level annotations are not available.}
    \label{fig:lymph}
    \vspace{-10pt}
\end{figure}

Deep learning methods have contributed much to the domain of medical image analysis, by solving image-/pixel-level classification and regression problems. 
However, whole-slide image (WSI) analysis remains to be challenging for deep learning, due to its uniqueness on the gigapixel size and the lack of pixel-level annotations, as illustrated in Figure \ref{fig:lymph}. 
The task of WSI classification, such as cancer classification, lymph node metastasis detection, and immunohistochemical scoring, is formulated as a weakly supervised learning problem that can be solved by multiple instance learning (MIL). 

Generally, all image patches from one WSI are regarded as instances to construct a bag, i.e., a collection of unlabeled instances, where all instances are negative if the WSI label is negative, and there exists at least one positive instance otherwise.  MIL methods aggregate instance-level information to obtain bag-level representations for bag-level classification. 
Existing deep learning based MIL methods try to localize the diagnostic relevant areas by applying feature embedding non-parametric pooling (e.g., mean/max pooling), self-attention~\cite{ilse2018attention,lu2021data}, non-local attention with Transformer~\cite{shao2021transmil}, etc, resulting in densely sampling for feature aggregation.

Current methods usually extract patch-level features for each instance based on a pretrained visual embedding, e.g., the training with ImageNet \cite{krizhevsky2012imagenet}, and self-supervised training using pathology data \cite{wang2021transpath,li2021dual}. 
Such operation can be performed once for all and then cached in the training stage to reduce the burden of feature extraction, which alleviates the computational intensity. 
%Therefore, the training of MIL models is not that computational intensive. 
To date, most weakly-supervised studies focus on extracting informative instance (patch) features and aggregating these features. 

However, during the inference stage, the drawback of having a heavy computational cost for extracting local patch features in an exhausted manner has often been neglected. 
With pretrained visual embeddings, e.g., ResNet50~\cite{he2016deep}, the feature extraction process usually requires dozens of minutes.
Such a limitation becomes evident in clinical applications,  as the MIL models require all instance features to perform the classification or regression. 
Therefore, it is necessary to develop a data-efficient method that can manage to avoid densely sampling all image patches in the high magnification and process informative features sparsely.

To this end, we aim to develop a fast observation strategy to avoid the aforementioned exhausting feature extraction. 
%A recent study demonstrated that it is possible to represent the WSI with sparsely sampled image patches in a tree structure~\cite{chen2021diagnose}. 
Motivated by the decision-making process of human pathologists who sparsely analyze field of views at different scales by switching objective magnification powers of the microscope, it would be inspiring to have a fast observation strategy that selectively observes diagnostic-relevant regions without having to analyze each part of the WSI at a high magnification. 

%For WSI classification task like lymph node metastasis detection (binary classification of WSI), especially, only a small portion of each positive slide contains tumour regions that correspond to slide-level labels. Without pixel-level or region-of-interest (ROI) -level annotation of these tiny regions, variants of multiple instance learning (MIL) approaches are usually adopted to model slide-level stratification as a weakly supervised learning problem. Existing deep learning based MIL methods try localize the diagnostic relevant areas by applying feature embedding non-parametric pooling (e.g., mean/max pooling), self-attention mechanisms~\cite{ilse2018attention,lu2021data}, non-local attention with transformer~\cite{shao2021transmil}, etc, resulting in a weighted navigation and observation strategy that traverses all areas in the WSI. 

In this paper, we model the diagnostic decision-making process as a Markov Decision Process (MDP), where a human-imitating agent first looks for regions of observation value by scanning at low magnification and then choose potential diagnostic-relevant regions to observe in depth. 
Based on the idea of super-resolution, the agent is designed to continuously accumulate knowledge of what the scene of high magnification may look like for an unobserved low-magnification region by browsing across multiple resolution levels, i.e., updating the low-level representation of unobserved areas with observed region feature pairs at different resolution levels. 
Consequently, a quick slide-level classification decision is made by repeating the aforementioned pattern to generate a short observation pathway. 
Inspired by AI for gaming \cite{lin2022juewu,ye2020towards,ye2020mastering}, 
we train an agent to solve such a path-decision problem which is inherently related to optimal path selection with deep reinforcement learning (DRL). 

To sum up, we make the following contributions: 
\begin{itemize}
    \item We develop RLogist, a DRL method for whole-slide image analysis, so as to avoid analyzing local image patches in a traditional exhaustive manner. 
    \item Inspired by the decision-making process of a human, we implement the cross-magnification information fusion by conditional feature super-resolution. Benefiting from the conditional modeling, the high-magnification features of unobserved region could be updated according to the sparsely observed low-magnification and high-magnification feature pairs.  
    \item We propose a well-defined environment for reinforcement learning algorithms, using the discrete action space to improve the model convergence performance by patching and the done-state reward to avoid local optima.
    \item Extensive experiments demonstrate that the performance of our RL agent is comparable to state-of-the-art methods while having a significantly shortened observation path. 
\end{itemize}

\section{Related Work}

We develop a fast observation strategy for WSI analysis using DRL models. In this section, we review recent efforts on WSI analysis with MIL framework, as well as relevant works on DRL methods in medical image analysis.

\subsection{WSI Analysis with Multiple Instance Learning}
Weakly supervised  learning with neural networks has been widely adopted for many classification tasks such as tumor detection, because the diagnostic areas that contributes to the slide-level label only occupies a relatively small proportion of the WSI \cite{campanella2019clinical}. 
Without detailed annotation of diagnostic-relevant regions, i.e., only slide-level labels are available, WSI classification is usually cast as a multiple instance learning (MIL) problem \cite{carbonneau2018multiple}. Recent works have shown that an MIL classifier trained on large weakly-labeled WSI datasets even generalizes better than a fully-supervised classifier trained on pixel-level-annotated small lab datasets \cite{bulten2022artificial,campanella2019clinical}. 

Conventional MIL models use handcrafted operators, such as mean-pooling or max-pooling to aggregate all feature embeddings of the patches in a whole slide \cite{pinheiro2015image, feng2017deep}. To improve the data efficiency and classification performance, variant MIL attention based methods assign more accurate contribution weight of each instance by introducing neural network based aggregators with trainable parameters \cite{hashimoto2020multi, zhao2020predicting, xie2020beyond}. 
Non-local attention mechanisms are introduced to model the relations of the instances in a dual-stream architecture with trainable distance measurement \cite{li2021dual}. 
Transformer is also adopted in MIL to explore morphological and spatial information among different instance \cite{shao2021transmil}. 
\citet{chen2021diagnose} proposed an aggregation operator with relevance-enhanced graph convolutional network to jointly analyze visual fields in a multi-scale manner for diagnostic predictions. 
In short, recent MIL methods mainly focus on assigning better contribution weights to patch instances by exploring and exploiting the correlated information between different instances or different resolution levels \cite{chen2022scaling}. Few of them focus on fast observation path planning. 
%In our experiments, we compare our work with these state-of-the-art MIL-based works. 

\subsection{DRL Methods in Medical Image Analysis}
Deep reinforcement learning (DRL) augments the reinforcement learning
framework, which learns a sequence of actions that maximizes the expected
reward, with the representative power of deep neural networks \cite{li2017deep, silver2017mastering, zhao2021augmenting}. Recent works have demonstrated the great potential of DRL in medicine and healthcare \cite{zhou2021deep}. DRL methods are widely applied in parametric medical image analysis tasks including landmark detection, object/lesion detection, registration, and view plane localization. \citet{alansary2019evaluating} evaluate different reinforcement learning agents with several training strategies for detecting anatomical landmarks in 3D images. \citet{qaiser2019learning} treat IHC scoring as a sequential learning task to learn discriminative features and select informative regions within a cropped image ($2048\times2048$) of a WSI. Note that although this method is designed for pathology image classification, it is not designed for whole-slide images.  Because DRL can handle the non-differential metrics, it is widely used to solve optimization problems where conventional methods fall apart \cite{zhou2021deep}. Such applications includes tuning hyperparameters for radiotherapy planning \cite{shen2018intelligent}, selecting the right image augmentation selection for image classification \cite{cheng2019adversarial, ye2020synthetic, yang2020deep, wang2020auto}, searching best neural architecture for segmentation \cite{furuta2019pixelrl, pineda2020active}, and avoiding poor images via a learned acquisition strategy \cite{zaech2020learning}. DRL methods are also used in miscellaneous applications, such as surgical gesture segmentation \cite{liu2018deep}, personalized mobile health intervention \cite{zhu2018group}, and computational model personalization \cite{abdi2020muscle, joos2020reinforcement}.

However, realizing the full potential of DRL for medical imaging requires solving several challenges that includes defining an appropriate reward function, data availability, dynamic environment and etc \cite{zhou2021deep, coronato2020reinforcement}. In addition, advanced DRL algorithms which have been proven to perform better and converge easier in many application fields, such as Proximal Policy Optimization (PPO) \cite{schulman2017proximal} and Soft Actor-Critic (SAC) \cite{haarnoja2018soft}, are still not widely used in computational pathology. In this paper, for the first time, we propose a generic deep reinforcement learning framework for observation and prediction strategy on WSIs with a precise definition of the state and action space, dealing well with dynamic environments and reward design issues.

\begin{figure*}[htbp]
    \centering
    \includegraphics[width=0.95\textwidth]{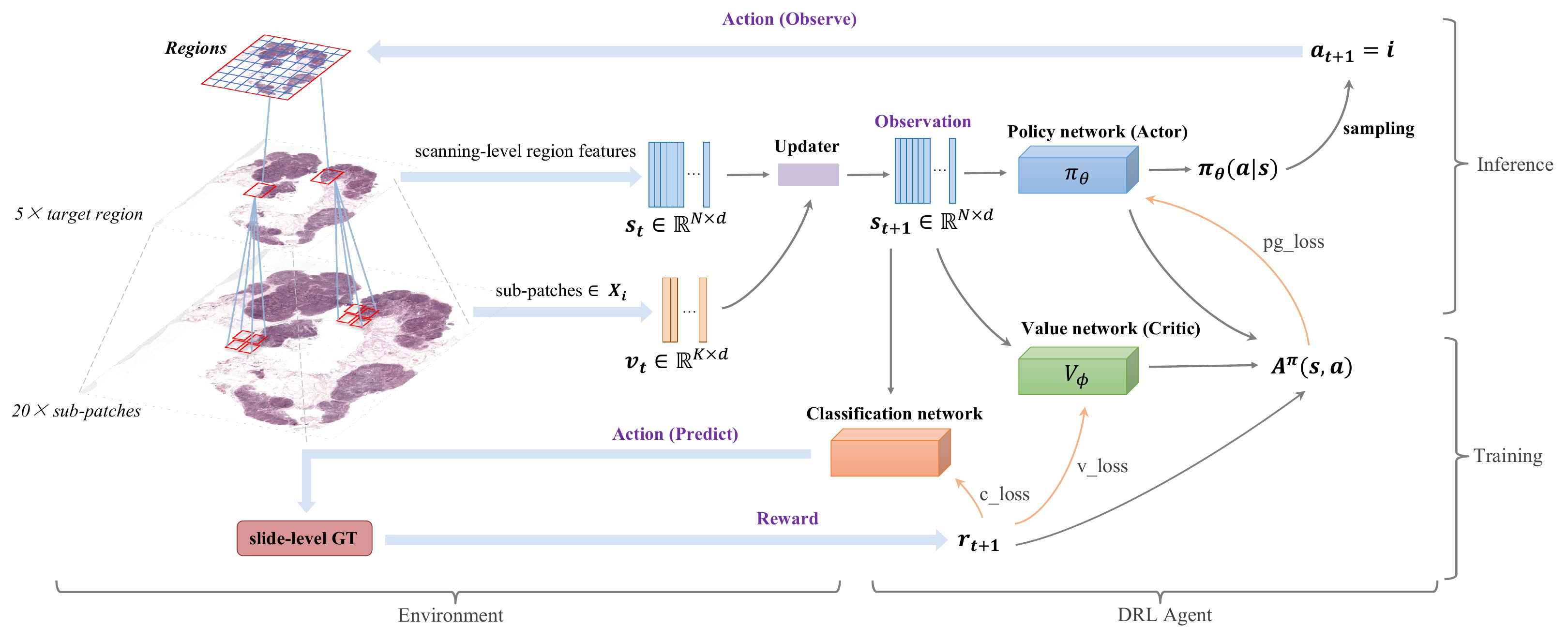}
     \vspace{-10pt}
    \caption{The architecture of RLogist. In the agent-environment loop, the blue arrows represent interaction with the environment, while the grey arrows represent the data pathway in the reinforcement learning model.}
    \label{fig:architecture}
    \vspace{-13pt}
\end{figure*}

\section{Method}

In this section, we depict the design of RLogist for fast observation strategy on whole-slide images. The overview architecture is shown in Figure \ref{fig:architecture}. We first introduce the formulation of the observation process on WSIs, which is modeled as a Markov Decision Process (MDP), then present the reinforcement learning architecture and concrete descriptions of the components in our proposed model.

\subsection{Problem Formulation}
Given a WSI $I$, the task is to predict the slide-level label by scanning the image at low magnification (global visual perception) and observing potential diagnostic-relevant regions in depth at high magnification (local visual perception). 
The objective is to find the shortest possible observation path,  while ensuring the slide-level prediction performance.

After tissue region segmentation, non-overlapping image patches with resolution $[l \times l ]$ of multiple target magnification $M$ are extracted from the WSI, annotated as $\{x_{i}^{M}\}_{i=1}^{N}\in\mathbb{R}^{N\times [l \times l]}$, where $N$ is the number of patches varying from slide to slide. With a feature extractor $f$, each patch instance $x_{i}$ can be described by a fixed $d$-dimensional vector representation $\bm{v}_{i} = f(x_{i}) \in \mathbb{R}^{1\times d}$. 
For image patches of the scanning-level magnification in a WSI, we refer to the collection of the extracted feature vectors as $\{\bm{v}_{i}\}_{i=1}^{N}\in\mathbb{R}^{N\times d}$. 
The feature vectors of higher magnification patches belonging to each scanning-level magnification patch $x_{i}$ are annotated as $\{\bm{v}_{j}^{i}\}_{j=1}^{K}\in\mathbb{R}^{K\times d}$, where $K$ is the number of higher magnification sub-patches. 
For example, an observation strategy with 5x scanning-level magnification and 20x observation-level magnification results in the sub-patch number $k = 16$.

Imitating the observation pathway of human pathologists on a whole-slide image, we model the decision-making process as a Markov decision process (MDP). 
At each time step $t$, the agent gets an observation $\{\bm{v}_{i}^{t}\}_{i=1}^{N}$ of the state $s_{t}$ by scanning at low magnification and then choose a potential diagnostic-relevant region $x_{i}$ to observe in depth as action $a_{t}$. 
After observing sub-patches in region $x_{i}$ to get local detailed features $\{\bm{v}_{j}^{i}\}_{j=1}^{K}$, the process responds at the next time step $t'$ by moving into a new state $s'$ with updated scanning-level representative features $\{\bm{v}_{i}^{t'}\}_{i=1}^{N}$. This step is more like a pathologist can infer unseen details through global visual perception and partial detailed field of views.
In certain observation steps, the agent generates a prediction of the slide-level classification label and gets a trajectory reward $R_{\tau}$, which is correlated with the cross-entropy of the output logit and the ground-truth. With a deep reinforcement model, we can train the agent to learn a parameterized policy $\pi_{\theta}$ optimizing the length of the observation path, i.e., to obtain a fast observation strategy. 

In our task, there are several key problems that need to be solved. First, how observation actions affect the state transition process, i.e., updating scanning-level representative features $\{\bm{v}_{i}^{t'}\}_{i=1}^{N}$. Second, how to handle the trade-off between the length of the observation path and the classification performance. We will elaborate our RLogist solution to these problems in the following subsections.

\subsection{RLogist: Reinforcement Learning for Observation}

In the proposed model, our key innovation is that the RL agent not only develops a discriminative capacity of diagnostic-relevant regions in the training stage, but also gradually improves the cognitive ability of unobserved regions by accumulating information about the observed regions in the inference stage. 
Based on the idea of super-resolution, the latter mechanism conditionally updates the scanning-level feature of unobserved regions $\{\bm{v}_{i}\}_{i=1}^{N'}$ with the low-high feature pair of the new-observed region, which can improve the accuracy of target region selection and accelerate the observation process significantly. We now describe each components of the RLogist in detail.

\subsubsection{Deep Reinforcement Learning Framework}

\begin{algorithm}[htbp]
\caption{The Agent-Environment Loop at Step $t$}
% \textcolor{red}{the algorithm name is strange}
\label{alg:agentenvloop}
\textbf{Input}: Current state $s_t$; \\
\textbf{Parameters}: Policy parameter $\theta_{t}$; State transition function $f(s,a)$ represented by local feature updater $f_{local}$ and unobserved region feature updater $f_{global}$;\\
\textbf{Output}: Next state $s_{t+1}$ and action $a_{t}$; \\
\begin{algorithmic}[1] %[1] enables line numbers
\STATE Sample $a_{t}$ from the distribution $A_{t} \sim \pi_{\theta_{t}}(\cdot | s_{t})$
\IF{training}
\STATE Update the policy $\pi_{\theta}$ with an RL algorithm
\ENDIF
\STATE Observe a region corresponding to $a_t$ in the WSI and get a set of low-level feature embedding $\{\bm{v}_{j}^{a_t}\}_{j=1}^{K}$
\STATE Update the scanning-level feature of region $a_t$: \\ $\bm{v}'_{a_t} = f_{local}(\{\bm{v}_{j}^{a_t}\}_{j=1}^{K})$
\FOR{scanning level region index $i$ \textbf{in} $\{1, 2, ... N\} \setminus a_t$}
\STATE The unobserved region: $\bm{v}'_i = f_{global}(\bm{v}_i, (\bm{v}_{a_t}, \bm{v}'_{a_t}))$
\ENDFOR
\STATE Transit to next state $s_{t+1}$ with ${\bm{v}'_i}$ and $\bm{v}'_{a_t}$
\end{algorithmic}
\end{algorithm}

Our target is to develop a fast observation strategy, and we achieve it by learning a parameterized policy $\pi_{\theta}(a|s)$ optimizing the classification performance within a limited number of observation steps. The terminology $\pi_{\theta}(a|s)$ is defined by:
\begin{equation}
   \pi_{\theta}(a|s) = P(A_t|S_t; \theta)
\end{equation}
where $A_t$ and $S_t$ denotes the set of possible actions and states at time step $t$.
Acting as the backbone of our proposed model, the stochastic policy network takes scanning-level feature embeddings $s_t$ as the input at time step $t$, and outputs the logits of each legal actions, i.e., target regions for the next observation move. Note that, the explored regions need not be re-observed, we mask out them by replacing the logits of these unavailable actions with large negative number (e.g., $-1 \times 10^8$) before sampling the action $a_t$. The process of an agent scanning the whole image at low-magnification is essentially to analyze multiple high-level regions in parallel with the policy network $\pi_{\theta}(a|s)$. After receiving the action $a_t$ to observe a new target region, the state transits to the next state $s_{t+1}$ by updating scanning-level feature with local feature updater $f_{local}$ and unobserved region feature updater $f_{global}$, which will be described in detail within the following subsection. Through this environment design, we can handle the problem of changing action spaces that the number of low-magnification instances extracted from different WSIs after patching may be different. The agent-environment loop is summarized in Algorithm \ref{alg:agentenvloop}.

The learning task is to optimize the parameters $\theta$ that 
maximize the overall performance $J(\pi_{\theta})$:
\begin{equation}
   J(\pi_{\theta})=\int_{\tau} P(\tau \mid \pi_{\theta}) R(\tau)=\underset{\tau \sim \pi_{\theta}}{\mathrm{E}}[R(\tau)]
\end{equation}
where $R(\tau)$ is the expected return of the trajectory $\tau$, associated with reward $r_t$. Given that only slide-level labels are available, we indirectly give the feedback (reward $r_t$) to the policy of observation path selection, using the classification results generated at each steps and the end of each trajectory. 

Here we discuss the formation of rewards in detail and give corresponding proofs, as it is critical to design rewards which are consistent with our ultimate target. Actually, the objective of finding a fast observation path can be modeled differently as these two optimization problems: 

\noindent(1) Jointly optimize the slide-level prediction performance and the length of the observation path; 
\begin{equation}
\label{eq3}
    \theta = \arg \underset{\rho} \min [L_c, \ell]
\end{equation}
(2) With a fixed length of the observation path, merely optimize the slide-level prediction performance
\begin{equation}
    \theta = \arg \underset{\rho, \ell_{0}} \min L_c
\end{equation}
where $L_c$ is the classification loss, and $\ell$ is the length of the observation path. For binary classification tasks, the cross entropy loss can be calculated as: 
\begin{equation}\label{CE_loss}
    L_c = -(y \log (\hat{y})+(1-y) \log (1-\hat{y}))
\end{equation}

For the former target, the agent may predict the slide-level label $\hat{y_t}$ at each time step $t$, and an instant reward is given by comparing $\hat{y_t}$ with the ground-truth $y_t$ with cross entropy. This instant reward design can also be used on the latter objective to accelerate the convergence. However, if the model receives a direct reward after interacting with each selected regions, the agent will eventually learn a greedy strategy that may converge to the local optimum. Under certain circumstances, observing a specific region first may not be the best choice for prediction at this time step, but can help the agent to gain a better understanding of the whole slide that benefits following region-selecting decisions, resulting in better performance after certain steps than greedy strategies. 

On common datasets, the trajectory length is relatively short for scanning at low-magnification (usually less than 100 time steps), making it easy to converge and feasible to just use a final reward at the done state when optimizing the policy. In the experiments, we only use the final reward of each trajectory to learning a more comprehensive strategy. 

With this well-defined environment, it is crucial to adopt an effective while efficient policy gradient algorithm for the entire reinforcement learning process to optimally leverage the reward. We use the Proximal Policy Optimization (PPO) in our method due to its computational efficiency and satisfactory performance. Here, we briefly introduce how the PPO algorithm optimize the policy for region selection in the training stage. PPO-clip updates the policy $\pi_{\theta}$ via maximizing the objective $L(s, a, \theta_{k}, \theta)$:
\begin{equation}
    \theta_{k+1} = \arg \max_{\theta} \underset{s, a \sim \pi_{\theta_{k}}} {\mathrm{E}}\left[L\left(s, a, \theta_{k}, \theta\right)\right]
\end{equation}
Based on the idea of importance sampling \cite{kloek1978bayesian}, the weight assigned to the current policy also depends on older policies. The probability ratio between previous and current policies is defined by:
\begin{equation}
    \gamma(\theta, \theta_{k}) = \frac{\pi_{\theta}(a \mid s)}{\pi_{\theta_{k}}(a \mid s)} 
\end{equation}
Then $L(s, a, \theta_{k}, \theta)$ is given by:
\begin{equation}\label{pg_loss}
    L = \min \left(\frac{\pi_{\theta}(a|s)}{\pi_{\theta_{k}}(a|s)} A^{\pi_{\theta_{k}}}(s, a), g\left(\epsilon, A^{\pi_{\theta_{k}}}(s, a)\right)\right)
\end{equation}
where 
\begin{equation}
    g(\epsilon, A)= \begin{cases}(1+\epsilon) A & A \geq 0 \\ (1-\epsilon) A & A<0\end{cases}
\end{equation}
$\epsilon$ is a relatively small hyperparameter to clip the objective, which roughly says how far away the new policy is allowed to go from the old.

Through optimizing the classification network with Eq. \ref{CE_loss} and policy network with Eq. \ref{pg_loss}, respectively, 
our RLogist model can obtain the remarkable path-selection and classification performance only using the slide-level weak labels. The training pipeline of the DRL models (policy network $\pi_{\theta}$ and critic network $V_{\phi}$) based on the actor-critic architecture and PPO algorithm is demonstrated in Algorithm \ref{alg:RLPipeline}. 
We compare the performance of PPO with classic policy gradient algorithms like REINFORCE \cite{luthans1999reinforce} in the experiments.

\begin{algorithm}[htbp]
\caption{DRL Training Pipeline}
\label{alg:RLPipeline}
\textbf{Input}: Initial policy parameter $\theta_{0}$; initial value function parameters $\phi_{0}$; \\
\textbf{Parameters}: Starting state distribution $\rho_0$; \\
\textbf{Output}: The trained policy network; \\
\begin{algorithmic}[1] %[1] enables line numbers
\FOR{$k=0,1,2,...$}
\FOR{$i=0,1,2,...$}
\STATE Run policy $\pi_{k} = \pi(\theta_{k})$ in the environment to generate an observation path $\tau_{i}\ \langle s, a, r, p, \rho_0 \rangle$.
\ENDFOR
\STATE Collect a set of trajectories $D_{k}=\{\tau_{i}\}$. 
\STATE Estimate the advantage $\hat{A_{t}}$ with current value function $V_{{\phi}_{k}}$ using Generalized Advantage Estimate.
\STATE Update the policy by maximizing PPO Objective: \\
$\theta_{k+1}=\arg \max _{\theta} \frac{1}{\left|\mathcal{D}_{k}\right| T} \sum_{\tau \in \mathcal{D}_{k}} \sum_{t=0}^{T} L\left(s, a, \theta_{k}, \theta\right)$
\STATE Compute Reward-to-go $\hat{R_t}$.
\STATE Update the value function with MSE loss:
$\phi_{k+1}=\arg \min _{\phi} \frac{1}{\left|\mathcal{D}_{k}\right| T} \sum_{\tau \in \mathcal{D}_{k}} \sum_{t=0}^{T}\left(V_{\phi}\left(s_{t}\right)-\hat{R}_{t}\right)^{2}$
\ENDFOR

\end{algorithmic}
\end{algorithm}
% $L\left(s, a, \theta_{k}, \theta\right)=\min \left(\frac{\pi_{\theta}(a \mid s)}{\pi_{\theta_{k}}(a \mid s)} A^{\pi_{\theta_{k}}}(s, a), \quad \operatorname{cip}\left(\frac{\pi_{\theta}(a \mid s)}{\pi_{\theta_{k}}(a \mid s)}, 1-\epsilon, 1+\epsilon\right) A^{\pi_{\theta_{k}}}(s, a)\right)$

% $\phi_{k+1}=\arg \min _{\phi} \frac{1}{\left|\mathcal{D}_{k}\right| T} \sum_{\tau \in \mathcal{D}_{k}} \sum_{t=0}^{T}\left(V_{\phi}\left(s_{t}\right)-\hat{R}_{t}\right)^{2}$

\subsubsection{Fast Convergence with Pretrained Classifier} 
At the end of an observation path, the slide-level label is predicted with the information of the observed regions by the classification network. However, training two networks from scratch simultaneously in an RL algorithm leads to convergence difficulties. During the cold start phase of model training, we apply a pretrained classifier to accelerate convergence.
We train CLAM classifiers~\cite{lu2021data} on the training set of corresponding datasets in different magnifications. We split the classification multilayer perceptrons (MLPs) from the models and used them as the patch classifiers at different magnifications.  By this means, the policy network is expected to focus on developing the path selection ability with a fixed classification network. 

\subsubsection{Unobserved Region Feature Update} 
In our environment, the state transition process is to extract the  high-power feature corresponding to the action, and then update the scanning-level representation with it. Firstly, the environment updates the scanning level feature of the currently-observed region, with the newly extracted high-power feature embedding of the sub-patches belonging to this region and the local feature updater $f_{local}$. $f_{local}$ can be implemented as a Vision Transformer \cite{dosovitskiy2020image} and pretrained by self-supervision, taking the sequence of feature embeddings of these sub-patches in the region $a_t$ as the input:
\begin{equation}
    \bm{v}'_{a_t} = f_{local}(\{\bm{v}_{j}^{a_t}\}_{j=1}^{K})
\end{equation}
where $\bm{v}'_{a_t}$ is the updated local scanning-level feature embedding. 

Then, with the pair of feature embedding $(\bm{v}_{a_t}, \bm{v}'_{a_t})$ for region $a_t$, the scanning-level representation of unobserved regions can be updated with $f_{global}$, implemented as
an MLP-head. This MLP-head takes the concatenated vector $[\bm{v}_i, \bm{v}_{a_t}, \bm{v}'_{a_t}]$ as the input:
\begin{equation}
   \hat{\bm{v}'_{i}} = f_{global}(\bm{v}_i, \bm{v}_{a_t}, \bm{v}'_{a_t})
\end{equation}
The objective of the MLP prediction is the local region feature directly generated from its corresponding high-magnification sub-patches. Therefore, the MLP-head can be pretrained with the ground-truth generated by $f_{local}$:
\begin{equation}
   \bm{v}'_{i} = f_{local}(\{\bm{v}_{j}^{i}\}_{j=1}^{K})
\end{equation}

\section{Experiments}

In our experiments, we use two publicly available WSI datasets, i.e., the CAMELYON16 lymph node WSI dataset \footnote{https://camelyon16.grand-challenge.org/},  and the TCGA-NSCLC lung cancer dataset \footnote{https://www.cancer.gov/tcga}, 
to demonstrate the effectiveness of our DRL fast observation strategy and the unobserved region feature update mechanism. 

\subsection{Dataset}
CAMELYON16 is a typical dataset for breast-cancer lymph-node-metastasis detection. CAMELYON16 consists of 270 annotated whole slides for training and another 129 slides as a held-out official test set. We directly use the officially divided training set and test set in our experiments. After segmentation and patching, the dataset yields roughly 0.25 million patches at 5× magnification and 3.2 million patches at 20× magnification with on average about 8,000 and 625 patches per WSI. 

The TCGA-NSCLC dataset consists of 1054 diagnostic WSIs from the TCGA NSCLC repository under the TCGA-LUSC and TCGA-LUAD projects. 10 low-quality slides without MPP and magnification information are discarded. We randomly split the dataset into 835 training slides and 209 testing slides with stratified sampling from 2 different TCGA projects. After pre-processing, a total of about 0.3 million patches at 5× magnification and 4.8 million patches at 20× magnification with on average about 350 and 5600 patches per WSI were obtained. 

For both the datasets mentioned above, every WSI is cropped into 224 × 224 patches without overlap and only slide-level labels are used to train our models.

\subsection{Experimental Setup}
We adopt the ImageNet pretrained ResNet-50 \cite{he2016deep} for the feature extraction network as $f$, which converts patches into 1024-dimensional features $\boldsymbol{v}$. For DRL, we implement policy gradient algorithm PPO2 roughly matching the details in stable-baselines3 \cite{stable-baselines3}. We also refer to the code-level optimizations presented in openai/baselines's PPO \cite{baselines}, implementation details in CleanRL \cite{huang2021cleanrl} and some of the tricks summarized by \citet{shengyi2022the37implementation}. We use the Adam optimizer with an annealing learning rate, which is initialized as $2.5 \times 10^{-4}$ and decays to 0 as the number of time steps increases, and $\epsilon=1 \times 10^{-5}$ to update the model weights during the training of our policy network. Cross-entropy loss is adopted to calculate the trajectory reward. We refer to the API design of OpenAI Gym \cite{aaaa} to implement our RL environment. For CLAM classifier, we follow the implementation details and experiment settings of \citet{lu2021data}. At the training stage, the CLAM pretrained classifier at low magnification (scanning level) is adopted to help initialize the policy network. All models in the experiment are implemented in PyTorch and trained until convergence on half NVIDIA Tesla T4 GPU. 

\begin{figure*}
    \centering
    \includegraphics[width=0.95\textwidth]{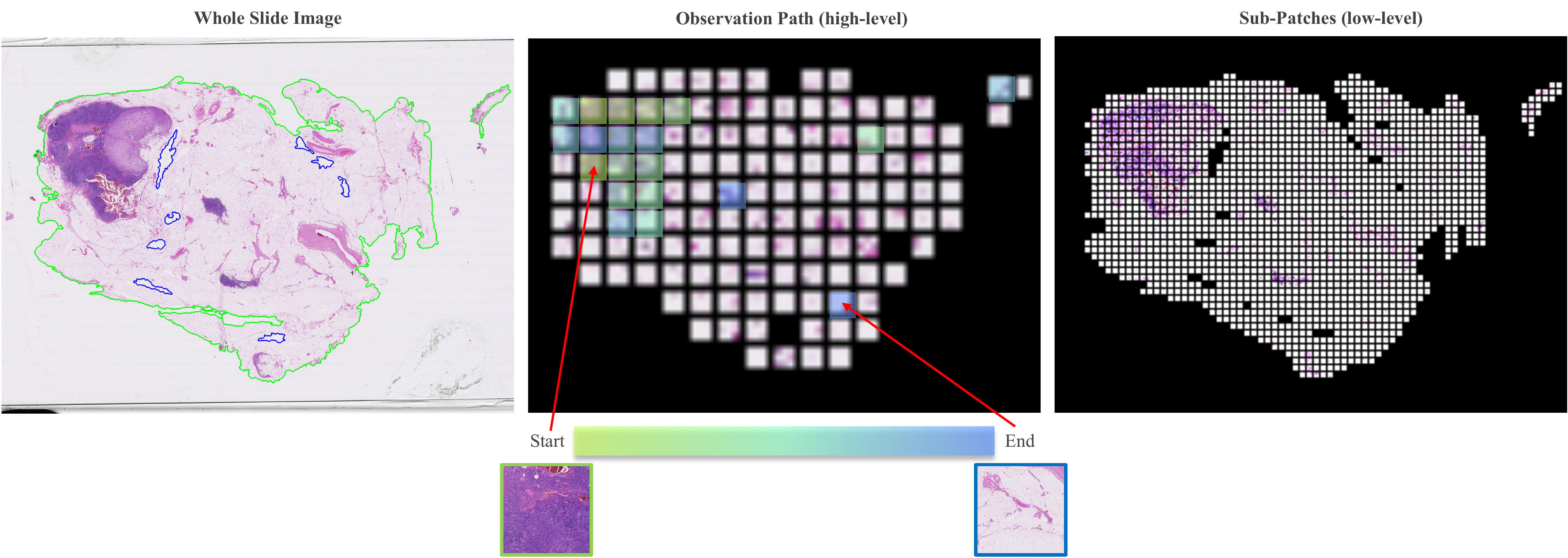}
    \vspace{-6pt}
    \caption{Visualized observation path of the RLogist on lymph node metathesis WSIs. The RLogist-0.2 starts the observation path from the region marked in green and ends in the blue region. About 20\% of the patches were covered to make slide-level label predictions. It also seems that less than 20 steps can complete diagnosis for this WSI.}
    \label{fig:vis}

    \vspace{-10pt}
\end{figure*}

\subsection{Slide-Level Classification Results}
\begin{table}[t!]
\begin{minipage}{0.48\textwidth}
\centering
\caption{Classification results on CAMELYON16 and TCGA-NSCLC. The RLogist-0.1, RLogist-0.2, and RLogist-0.5 denote the models trained to observe different time steps before making a final prediction, corresponding to different proportions of the regions in WSI. For fairness, we do not compare our method with other methods that learn the high-level cross-instance information fusion, which may achieve better classification performance.~{\footnote{As one of the reviewers suggested, we report the results of TransMIL based on our implementation, and the results are slightly different from the originally reported results.}}}
\begin{tabular}{lcccc}
\toprule
  \multirow{2}{*}{Method} & \multicolumn{2}{l}{CAMELYON16} & \multicolumn{2}{l}{TCGA-NSCLC} \\
\cline{2-3} \cline{4-5} & \multicolumn{1}{c}{Acc.} & \multicolumn{1}{c}{AUC} & \multicolumn{1}{c}{Acc.} & \multicolumn{1}{c}{AUC} \\
\midrule
Mean-pooling            &0.6389 &0.4647 &0.7282 &0.8401               \\
Max-pooling             &0.8062 &0.8569 &\textbf{0.8593} &\textbf{0.9463} \\
ABMIL                   &0.8682 &0.8760 &0.7719 &0.8656               \\
DSMIL                   &0.7985 &0.8179 &0.8058 &0.8925               \\
TransMIL                &0.8449 &0.8769 &0.8565 &0.9303          
  \\
CLAM-SB                 &\textbf{0.8760} &0.8809 &0.8180 &0.8818               \\
% TransMIL                &0.8837 &0.9309 &0.8835 &0.9603               \\
\textbf{RLogist-0.1}       &0.8242 &0.8287 & 0.8284 &0.8920         \\
\textbf{RLogist-0.2}       &0.8624 &0.8794 & 0.8389 &0.9026         \\
\textbf{RLogist-0.5}       &0.8685 &\textbf{0.8874} &0.8405 &0.9031 \\ \bottomrule
\end{tabular}
\label{table1}
\end{minipage}
\end{table}

We evaluate and compare our method to a collection of baselines including (1) standard MIL algorithm using traditional aggregators such as mean-pooling or max-pooling and (2) recent state-of-the-art deep MIL models, on the tasks of WSI classification. We have trained 3 DRL agents that observe different proportions 10\%, 20\%, 50\% of the regions in each WSI , i.e. with fixed length of the observation paths, before making a final slide-level prediction, respectively. 

As shown in Table \ref{table1}, the performance of our RLogist is comparable to state-of-the-art methods (ABMIL \cite{ilse2018attention}, DSMIL \cite{li2021dual}, TransMIL~\cite{shao2021transmil}, and CLAM-SB \cite{lu2021data}), while having a significantly shortened observation path. RLlogist-0.1 represents the agent only observes 10\% of the regions in the WSI before generating slide-level classification results (0.2 and 0.5, respectively). 

We illustrate our fast observation strategy by comparing the run time of different methods in the inference stage. Experiments are performed on the CAMELYON16 test set, using the ImageNet pretrained ResNet-50 to extract 1024-dimensional features. The average inference run time for each WSI is reported in Table \ref{tabletime}. Note that the run time here includes the time consumption of WSI data preprocessing (segmentation and patching), feature extraction, and classification. As is shown, our RLogist reduces the burden of feature extraction significantly with a short observation path.

\begin{table}[t!]
\centering
\caption{Average inference time per WSI for slide-level classification on the test set, using half NVIDIA Tesla T4 GPU. Here "half" indicates the half computing power provided by a virtual GPU, which consumes much more time on feature extraction, compared to using a more powerful GPU, e.g., Tesla V100/A100.}
\begin{tabular}{lc}
\toprule
Method & Average Runtime / \SI{}{min} \\
\midrule
Mean/Max-pooling        &39.91             \\
CLAM-SB                 &40.89             \\
\textbf{RLogist-0.2}    &\textbf{9.74}     \\ \bottomrule
\end{tabular}
\label{tabletime}
\vspace{-8pt}
\end{table}

As is shown in Figure \ref{fig:vis}, the RLogist has observed most of the diagnostic-relevant (positive) regions in the first 10 steps, which are marked in green. For the vast majority of metastatic cases, positive areas are detected within 20 steps.

\subsection{Ablation Study}
To further determine the contributions of the DRL observation strategy and the feature update mechanism, we have conducted a series of ablation studies. All ablation study experiments are based on the CAMELYON16 dataset, for the positive slides in this dataset are highly unbalanced (a large portion of negative patches in a positive WSI) and the effect of fast observation strategy may be reflected better.

\subsubsection{Observation Strategy with DRL}
The observation policy is trained to correctly select regions that are either directly diagnostic-relevant or helpful for subsequent observation moves. Here, we compared the effect of a random observation strategy and the policies trained with our proposed method using REINFORCE and PPO algorithm.
For the random strategy, we randomly sample a proportion of regions from the WSI (consistent with the time steps of the RL agent), and predict the slide-level labels with CLAM using high-magnification feature embeddings of the sub-patches in these regions. The agent is trained to observe on average 20\% of the WSI foreground (RLogist-0.2). The classification and convergence results are shown in Table \ref{table2}. 

Compared with the random strategy, it can be seen that both REINFORCE and PPO algorithm can improve the classification performance, and PPO is more effective in diagnosis prediction and much easier to converge.

\begin{table}[t]
\centering
\caption{Ablations with different observation strategies.}
\begin{tabular}{lcc}
\toprule
Observation Strategy & AUC    & \multicolumn{1}{l}{Convergence Episodes}\\
\midrule
Random (baseline)    & 0.7870 & /                        \\
REINFORCE            & 0.8364 & 142k                     \\
PPO                  & \textbf{0.8794} & \textbf{25k}    \\ 
\bottomrule
\end{tabular}
\label{table2}
\vspace{-12pt}
\end{table}

\subsubsection{Feature Update Mechanism}
Here we explore actual improvements for feature updaters $f_{local}$ and $f_{global}$. The performance of the model under different configurations is shown in Table \ref{table:feature}. Compared with the model without feature updater, it can be seen that both feature update mechanisms can enhance the classification performance, up to 0.53\% for $f_{local}$ and 2.74\% for $f_{local} + f_{global}$. 

\begin{table}
\centering
\caption{Ablations with different feature updaters.}
\begin{tabular}{lcc}
\toprule
Feature Updater & AUC    &\\
\midrule
Fixed feature (baseline)    & 0.8514       \\
$f_{local}$ & 0.8559             \\
$f_{local} + f_{global}$ & \textbf{0.8794}    \\ 
\bottomrule
\end{tabular}
\label{table:feature}
\vspace{-10pt}
\end{table}

% \section{Discussion} 

\section{Conclusions and Future Work}

We present a deep reinforcement learning method, named RLogist, for whole-slide image analysis. RLogist simulates a pathologist's decision-making process and can quickly complete the reading path navigation through the fusion of cross-magnification image information.
We design and implement the environment for RL training with discretized action space and conditional feature super-resolution mechanism. 
Competitive classification results have been achieved on two widely-used benchmark datasets, demonstrating the effectiveness and feasibility of our method. \
Notably, our method can significantly reduce the observation scale of image patches without losing much classification performance. Although recent AI-based screening/diagnostic algorithms have achieved satisfactory results on benchmark datasets, these data-driven methods, including ours, should be used with caution before being trained with large-scale clean data and extensive evaluated with multi-center clinical data. 

In the future, we would like to construct a better RL environment with fused cross-instance observation space using Transformers \cite{vaswani2017attention}, so as to enhance the feature representation. 
Moreover, it would be more reasonable to plan the observation path with variable length for each WSI as indicated in Eq.~\ref{eq3} and Fig.~\ref{fig:vis}. 
More sophisticated RL tricks should be used to avoid learning the wrong shortcuts.

%\bibliographystyle{aaai23}
%\bibliography{sample-base}
\bibliography{aaai23}

\end{document}